\documentclass[a4paper,UKenglish]{lipics-v2016}
 
\usepackage{color}
\usepackage{amssymb}
\usepackage{amsmath}
\usepackage[ruled,vlined]{algorithm2e}
\newtheorem{obs}{{\bf Observation}}

\usepackage{enumerate}
\usepackage{minipage}

\title{A Get-Together for Deaf and Dumb Robots in Three dimensional Space}

\author[1]{Subhash Bhagat}
\author[2]{Sruti Gan Chaudhuri}
\author[1]{Krishnendu Mukhopadhyaya}
\affil[1]{ACM Unit, Indian Statistical Institute,
   Kolkata, India.
  \texttt{sbhagat\_r@isical.ac.in, krishnendu@isical.ac.in}}
\affil[2]{Department of Information Technology, Jadavpur University, 
West Bengal, India.
 \texttt{srutiganc@it.jusl.ac.in}}
\authorrunning{S. Bhagat, S. Gan Chaudhuri and K. Mukhopadhyaya} 

\Copyright{Subhash Bhagat, Sruti Gan Chaudhuri and Krishnendu Mukhopadhyaya}

\subjclass{F.2.2  Geometrical problems and computations, 
I.2.11 Multi-robot systems, I.3.5 Geometric algorithms, languages, and systems}
\keywords{ Gathering, asynchronous, oblivious, polygonal obstacle, Swarm robots.}

\EventEditors{John Q. Open and Joan R. Acces}
\EventNoEds{2}
\EventLongTitle{42nd Conference on Very Important Topics (CVIT 2016)}
\EventShortTitle{CVIT 2016}
\EventAcronym{CVIT}
\EventYear{2016}
\EventDate{December 24--27, 2016}
\EventLocation{Little Whinging, United Kingdom}
\EventLogo{}
\SeriesVolume{42}
\ArticleNo{23}

\begin{document}

\maketitle

\begin{abstract} 
This paper proposes a strategy for a group of deaf and dumb robots, carrying clocks from different countries, to meet at a geographical location which is not fixed in advanced. The robots act independently. They can observe others, compute some locations and walk towards those locations. They can only get a snapshot of  the locations of other robots but can not detect whether they are static or in motion. The robots are forgetful; once they have completed their motion they forget their previous locations and observations. Again they decide new destinations to move to. Eventually all the robots compute the same destination and meet there. There exists no global positioning system. As they stand, they agree on up and down directions. However, as they do not have any compass, the other directions are not agreed upon. They also do not agree on  the clockwise direction. For determining a strategy, we imagine the robots to be points on a three dimensional plane where all the 
robots are mutually visible to each other always. The strategy we propose has to be obeyed by all the robots independently with respect to their own clock and compass. Initially the robots start from distinct locations. Some dead robots may be present in the system or some may die any time before or after the get together. However, the live robots are not aware of the presence of these dead robots.

{\bf Keyword:} 
Gathering, Asynchronous, Oblivious, 3 Dimensional plane, Swarm robots, Crash faults.

\end{abstract}

\section{Introduction}

An interesting branch of research in robotics is the study of  multi-robot systems, popularly known as {\it robot swarm}. A $robot$ $swarm$ is a collection of small autonomous, memoryless, communication-less, homogeneous, indistinguishable,
  inexpensive mobile robots working cooperatively to achieve some goal. 
Although large in numbers, collectively this swarm of robots is less expensive than a big robot. 
Increasing or decreasing the number of robots in this system involves very simple hardware or software modifications and thus provides good 
scalability. Moreover, having similar capability, if some robots fail, others can manage to execute the
work. This feature makes the system to be more resilient to malfunction.
 The fields of application for such a distributed system of robots are also versatile. One can use the robots to search for persons in a hazardous environment
\cite{sugihara1990distributed}, which typically include disaster hit areas. The robots can even work together to build a complex 3D structure \cite{michael2011cooperative}.
Other applications include mining in hazardous areas, agricultural tasks like foraging etc. Multi-robots systems are also used in defense. 
A large number of robots can act as an autonomous army. The U.S. Navy has created a swarm of boats which can track an enemy boat, surround it and then destroy 
it \cite{steinberg2006intelligent}. {\it Gathering or Homing} \cite{Santoro2012} or {\it Get-together}, (i.e., collecting the robots to a point not defined in advance) is a fundamental coordination problem for a 
group of mobile robots. This paper proposes an algorithm for a get-together of the multiple mobile robots deployed in a three dimensional plane. 

\subsection{Framework}
The distributed model \cite{Santoro2012} for a swarm of robots or multi robot system, represents the mobile entities by distinct points located in the region of deployment. Most of the existing literature deals with deployment in the Euclidean plane. This paper considers the {\it three dimensional} Euclidean plane. The robots
are anonymous, indistinguishable, having no
explicit communication through wired or wireless medium. As they stand, they agree on up and down directions. However, origin, axes, clockwise direction and unit distance are not same for the robots. 
 Each robot has sensing capability, by {\em vision}, which enables it to determine
 the positions (on its own coordinate system) of  the other robots.
The robots operate by executing {\em Look-Compute-Move} cycles asynchronously. 
All robots execute the same algorithm.
The robots are oblivious, i.e., at the beginning of each cycle, they forget
their past observations and computations. The robots execute the cycles asynchronously where the  robots may not start or complete cycles together. 
The vision enables the robots to communicate and coordinate their actions by sensing their relative positions. Otherwise, the robots are silent and have no explicit message passing. Some of the robots may be faulty in the sense that suddenly they can stop working forever.
   These restrictions enable the robots to be deployed in extremely harsh
environments where communication is not possible, e.g., an underwater
deployment or a military scenario where wired or wireless communications are
impossible or can be obstructed or erroneous.

\subsection{Contribution of This Paper}
{\it Gathering or homing} is one of the most visited problems \cite{Bouzid2013,Cieliebak2002,Cieliebak2003,Cieliebak2012,Degener2011,Dieudonne2012,Flocchini2005,Gordon2008,Izumi2007,Katayama2007,Prencipe2007} in the domain of the multi-robot 
systems. One of the goals of these investigations has always been to find out the minimum capabilities which  the robots must have to be gathered in finite time \footnote{there is a variation of gathering such as, convergence \cite{Cohen2005}; where the robots come as close as possible but do not gather at a single point. However, in this paper we only consider gathering at a single point.}. Deterministic gathering of $n>2$ asynchronous robots is impossible without the assumption on multiplicity detection (the robots can detect a point consists of multiple robots) or common agreement in coordinate systems or remembering the past \cite{Prencipe2007}. Flocchini et al. \cite{Flocchini2013}, have reported an algorithm for gathering two robots using constant number of memory bits. Flocchini et al. \cite{Flocchini2005}, have shown 
that gathering is possible if the robots have agreement in direction and orientation of both the axes, even 
when the robots can observe limited regions of certain radius, around themselves. If the robots agree on direction and orientation, Gathering is possible even if the robots have different visibility radii \cite{icdcit2015}. Agreement in coordinate system is an important parameter for gathering. If the robots agree only on orientation or chirality, i.e., they have common clockwise direction but no common direction of $X$ axis, gathering is not possible for two robots. 
Many researchers \cite{agmon2006fault,Defago2006,Bouzid2013,Bhagat201650} have considered errors in the system in various aspects, e.g., the robots can behave arbitrarily without properly following the algorithm or they can cease to work. 
Recently it has been shown that gathering of opaque robots is possible in two dimensional plane with only agreement in one axis and without any other assumption, even in the presence of faulty robots \cite{Bhagat201650}. This paper extends the result in \cite{Bhagat201650} to work for 3D. 
Forming of arbitrary pattern by the robots in 3D \cite{DBLP:journals/corr/YamauchiUY15}, forming a plane by the robots \cite{DBLP:conf/wdag/YamauchiUKY15} are some examples of the reported results on 3D. However, these results consider that the robots execute {\it look-compute-move} cycle synchronously. Gathering of multiple robots in a 3D plane is not yet reported. {\it In this paper we prove that one axis agreement is enough for gathering asynchronous, oblivious robots in three dimensional plane and propose a collision free gathering algorithm for such robots. We also consider that the robots may stop working forever at any point of time before or after meeting at a point.}


\section{Model and Terminologies}

The robots are represented as points in a three-dimensional plane. They are able to move freely on the plane.
Let $\mathcal{R}=\{r_1, r_2,\ldots, r_n\}$ 
denote the set of $n$ homogeneous, indistinguishable mobile robots. This robots follow the $CORDA$ model \cite{Santoro2012}
with some additional features. At any point of time, a mobile robot is either active or inactive(idle). The activation scheduling of the robots is asynchronous and 
independent from the others. Each 
active robot executes cycles consisting of  a sequence of states, namely {\it look-compute-move}, repeatedly in asynchrony. 
The duration of each operation and the delay 
between the intra-sequence operations are finite but  unpredictable. 
In {\it Look} state, a robot observes its surrounding in all directions and spots the positions of the other robot in
its local coordinate system to form a local view of the world. The robots do not share any common global coordinate system.
Each robot has its own local Cartesian coordinate system the origin of which is at the point in the space occupied by that robot. 
The robots agree only on one axis, the $Z$ axis. The orientation of the other two axes may be different for different robots. The robots are physically transparent, i.e., they not obstruct the visibility of
other robots. In {\it Compute} state, a robot, using the input received in the {\it Look} state, computes its destination point. Finally in {\it Move} state,
it moves to the computed destination. A robot may stop before reaching its destination and start a fresh computational cycle. 
However, to guarantee the finite time termination, it is assumed that 
whenever a robot moves, it travels at least a finite minimum distance $\delta > 0$ towards
its destination. The value of $\delta$ is not known to the robots. 
Due to asynchrony, a robot may observe other robots in motion. However, it can not detect
their motion. It only traces the locations of the mobile robots at the time of its observation. This implies that the computations of the robots may be based on locations which are no longer true. The robots do not remember the data computed in any of the previously completed cycles i.e., they are oblivious. The robots do not pass massages. 
Initially the robots are stationary. Multiple robots can occupy a single point.
However, it is assumed that the robots can not detect the presence of multiple robots at the same point i.e., they do not have the capability of $multiplicity$ detection.
Thus, the multiple occurrences of the robots at a single point is counted by the robots as a single occurrence.  
The robots may become faulty at any time during the execution of the algorithm. This paper considers \textit{crash
faults} where faulty robots stop executing cycles permanently. However, a faulty robot physically remains in the system without doing any action. The robots do not have the capability to distinguish the faulty robots. A crash fault model is denoted by $(n, f)$, where  at most $f < n$ robots can become faulty during the execution of the algorithm.

\begin{itemize}
 \item \textit{Configuration of the robots:} Let $r_i(t)$ denote the position occupied by the robot $r_i$ at time $t$. By a
configuration $\mathcal C(t) = \{r_1(t),\cdots, r_m (t)\}$, $m\le n$, we denote the set of positions occupied by the
robots in $\mathcal R$ at time $t$ (occurrence of multiple robots at a single point is counted once). 
Let $\widetilde{\mathcal C}$ denote the set of all such robot configurations.
\item The line segment joining two points $r_i(t)$ and $r_j(t)$ is denoted by $\overline{r_i(t)r_j(t)}$ and its length is denoted by $|\overline{r_i(t)r_j(t)}|$.
\item The positive direction of $Z$-axis is called as \textit{upward} direction and the negative direction of $Z$-axis is called as \textit{downward} direction. 
The directions which are perpendicular to the $Z$ axis will be called \textit{horizontal} directions; all other directions are \textit{non-horizontal}.
Planes, having $Z$ axis as their normal, are drawn through the robot positions in $\mathcal C(t)$. An ordering among these planes can be obtained according to their positions 
along the $Z$ axis from topmost downwards. Let $PL_i(\mathcal C(t))$ denote the $i^{th}$ plane from the top, 
containing the points in $\mathcal C(t)$ (Figure \ref{Theo-case-111}(a)). We define,

\begin{center}
 $\mathcal P(\mathcal C(t))=(PL_1(\mathcal C(t)), PL_2(\mathcal C(t)),\ldots, PL_k(\mathcal C(t)))$, $k\le n$.
\end{center}
 
Let $RPL_i(\mathcal C(t))$ denote the set of robot positions on the plane $PL_i(\mathcal C(t))$. We define,
\begin{center}
 $\mathcal RP(\mathcal C(t))=(RPL_1(\mathcal C(t)), RPL_2(\mathcal C(t)),\ldots, RPL_k(\mathcal C(t)))$.\\
\end{center}
\vspace{.5cm}
 The number of robot positions on $PL_i(\mathcal C(t))$ is denoted by $|RPL_i(\mathcal C(t))|$. 
 \vspace{.5cm}

   \begin{figure}[h]
\begin{center}
 \includegraphics[scale =1]{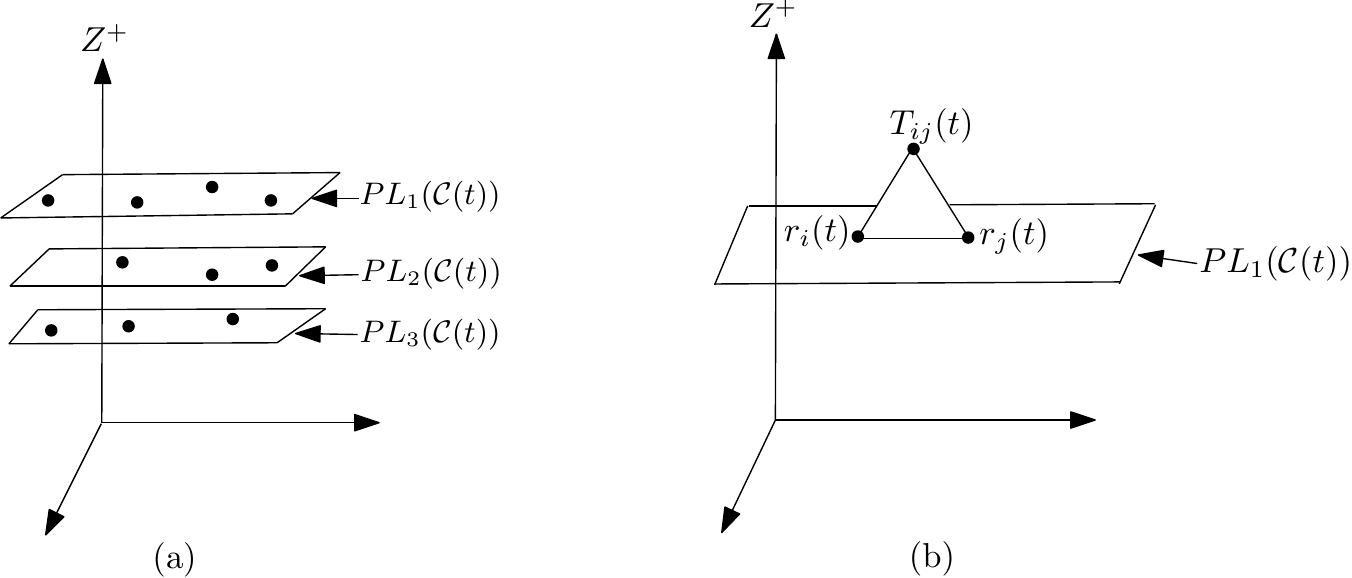}
 \end{center}
 \caption{An example of   $\mathcal P(\mathcal C(t))$ and $\triangle r_iT_{ij}r_j(t)$}
 \label{Theo-case-111} 
\end{figure}
 
 
\item
An equivalence relation $\prec$ is defined on $\widetilde{\mathcal C}$ as follows:
$\forall$ $ \mathcal C, \mathcal C' \in \widetilde{\mathcal C} $, 
$ \mathcal C\prec  \mathcal C'$ iff $|RPL_1(\mathcal C)|=|RPL_1(\mathcal C')|=1$ or $|RPL_1(\mathcal C)|=|RPL_1(\mathcal C')|=2$ or both $|RPL_1(\mathcal C)|$ and 
$|RPL_1(\mathcal C')|$ are greater than 2. 
This relation yields following three equivalence classes: (i) $\widetilde{\mathcal C_1}=\{\mathcal C\in \widetilde{\mathcal C}: |RPL_1(\mathcal C)|=1\}$  (ii) $\widetilde{\mathcal C_2}=\{\mathcal C\in \widetilde{\mathcal C}: |RPL_1(\mathcal C)|=2\}$ and (iii) $\widetilde{\mathcal C}_{>2}=\{\mathcal C\in \widetilde{\mathcal C}:
|RPL_1(\mathcal C)|>2\}$. 
 \item  For a plane $PL_m(\mathcal C(t))$ with $|RPL_m(\mathcal C(t))|\ge 2$, consider two points $r_i(t)$ and $r_j(t)$ on it. Let $\triangle r_iT_{ij}r_j(t)$ denote the equilateral triangle
 on $\overline{r_i(t)r_j(t)}$ and having the side length $|\overline{r_i(t)r_j(t)}|$ such that the direction of the perpendicular bisector of $\overline{r_i(t)r_j(t)}|$ is parallel to the $Z$ axis and $T_{ij}(t)$, the pick of the triangle, is on the upward direction (Figure \ref{Theo-case-111}(b)). 

 \item Consider $PL_1(\mathcal C(t))$ with $|RPL_i(\mathcal C(t))|>2$. If the robot positions on $PL_1(\mathcal C(t))$ are co-circular, let $S(PL_1(\mathcal C(t)))$ denote the circle  of the co-circularity. Otherwise, $S(PL_1(\mathcal C(t)))$ denotes the minimum enclosing circle of the robot positions on $PL_1(\mathcal C(t))$. The set 
 of robot positions on $S(PL_1(\mathcal C(t)))$ is denoted by $CS(PL_1(\mathcal C(t)))$. Let $W(S(PL_1(\mathcal C(t))), Z, 45^o)$ denote the right circular cone 
 with $S(PL_1(\mathcal C(t)))$ as the base,  axis of the cone parallel to $Z$ axis, semi-vertical angle equals to $45^o$ and vertex $V(PL_1(\mathcal C(t)))$ of the cone is on the upward direction 
 (Figure \ref{Theo-case-1112}).

   \begin{figure}[h]
\begin{center}
 \includegraphics[scale =.5]{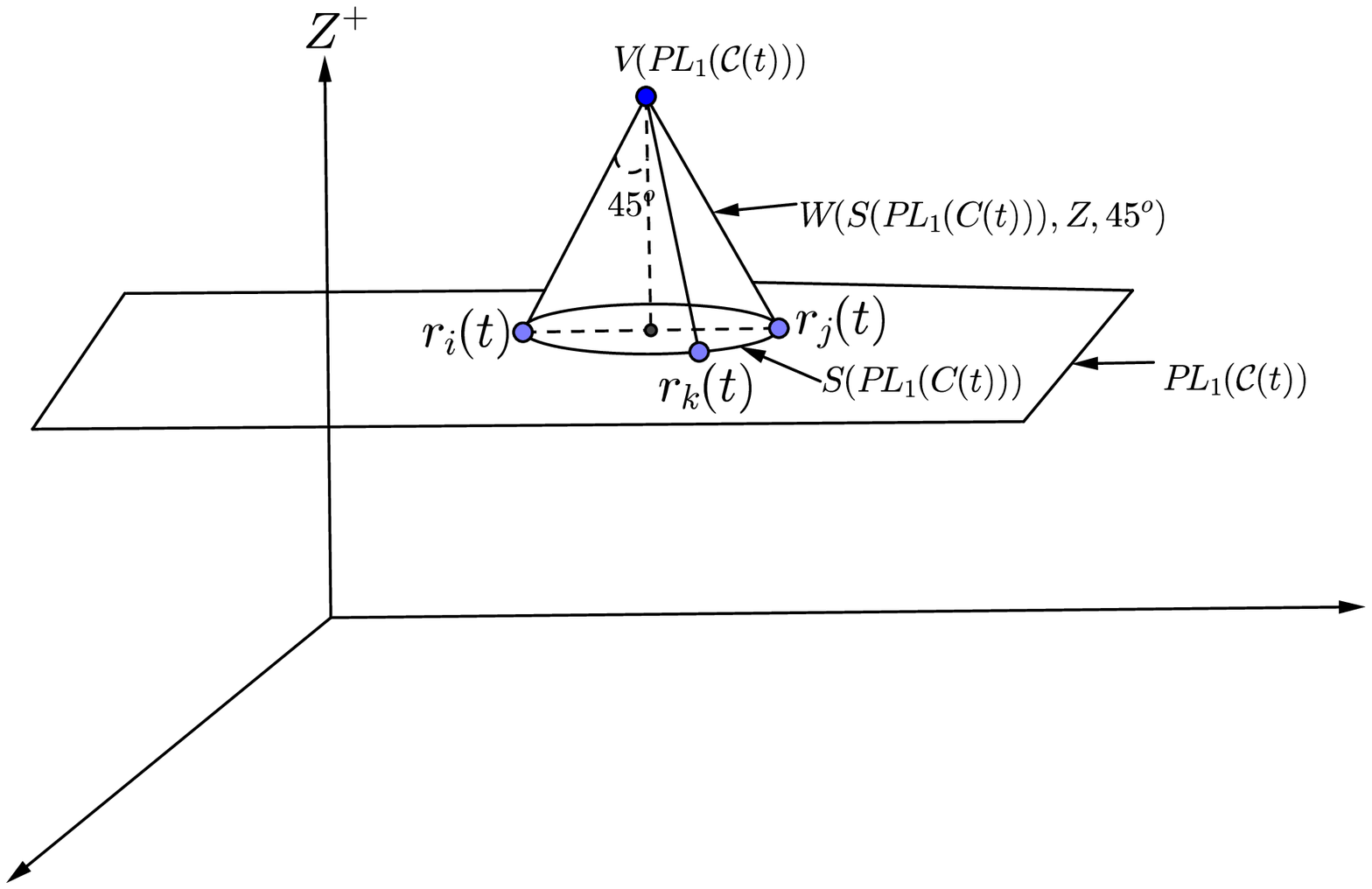}
 \end{center}
 \caption{An example of $W(S(PL_1(\mathcal C(t))), Z, 45^o)$}
 \vspace{-.7cm}
 \label{Theo-case-1112} 
\end{figure}

 
 \end{itemize}

 \begin{itemize}
 \item $\mathbf{CheckLevel3D():}$ This is a function which takes $\mathcal P(\mathcal C(t))$ and $r_i(t)$ as arguments. 
 It returns the value $k$ such that $r_i(t)$ is a point on the plane $PL_k(\mathcal C(t))$. 
 
 \item $\mathbf{ComputeCircle3D():}$ This function takes a set of points $\mathcal A$ lying on a plane as arguments. 
 This returns the circle passing through the points in $\mathcal A$ if the points are co-circular. Otherwise, it returns the minimum enclosing circle of the points in $\mathcal A$. It also returns the points of $\mathcal A$ which lie on the circumference of the computed circle.
\item $\mathbf{ComputeConeVertex():}$ This function takes the set of points on the circumference of  a circle $S(PL_i(\mathcal C(t)))$ as argument and computes  the cone
$W(S(PL_i(\mathcal C(t))), Z, 45^o)$. It returns  $V(PL_i(\mathcal C(t)))$, the vertex of the cone. 
%

\item $\mathbf{ComputeTrianglePeak3D():}$ This function takes two robot positions, $p_i(t)$ and $p_j(t)$ lying on the  plane $PL_m(\mathcal C(t))$ for some $m$, as argument. It computes the equilateral triangle $\triangle{p_iT_{ij}p_j}(t)$  
 and returns the point $T_{ij}$.
\item $\mathbf{ClosestPoint():}$ This is a function which takes a robot $r_i(t)$ and a set of points $\mathcal A$ as arguments. It returns the
closest position in $\mathcal A$ from $r_i(t)$ (tie, if any, is broken arbitrarily).
 
\end{itemize}

These functions are use to describe our proposed algorithm $\mathbf{Gathering3D}$. This algorithms is formally presented in section \ref{algo3d}.
The non-faulty robots in the system execute the algorithm $\mathbf{Gathering3D()}$ independently. During the execution of this algorithm, all the robots are allowed 
to perform their respective actions simultaneously i.e., the algorithm is { \it wait-free}.  

\section{Algorithm}
\label{algo3d} 

This section describes our gathering algorithm for a set of mobile robots $\mathcal R$. If it is possible to define a unique point which remains intact under 
 the motion of the robots, this point can be used as the point of gathering. Since the robots are oblivious, the movements of the robots towards gathering point
 should be planned carefully so that this point does not change. If such a point is not available in the system, the current configuration is changed  
 by the planned movements of the robots in finite time into the one 
 in which such a point is possible to define. Due to asynchrony, during the whole execution of the algorithm, it may be possible that different such points are discovered  by the robots. However, for gathering, the robots  must agree on a single point after a finite time.
 The different scenarios and corresponding solution strategies are as follows:\\ 

\begin{itemize}
 \item {\bf Case 1: $\mathcal C(t) \in \widetilde{\mathcal C_1}$:}
 The plane $ PL_1(\mathcal C(t))$ contains single robot position, say $r_i(t)$ (there may be multiple robots at this position but the robots can count all the appearances as the single one). The robots on $ PL_1(\mathcal C(t))$ do not change position. Rest of the robots move towards $r_i(t)$  
 along  the line segment joining their position to $r_i(t)$. 
 \begin{figure}[h]
\begin{center}
 \includegraphics[scale =.7]{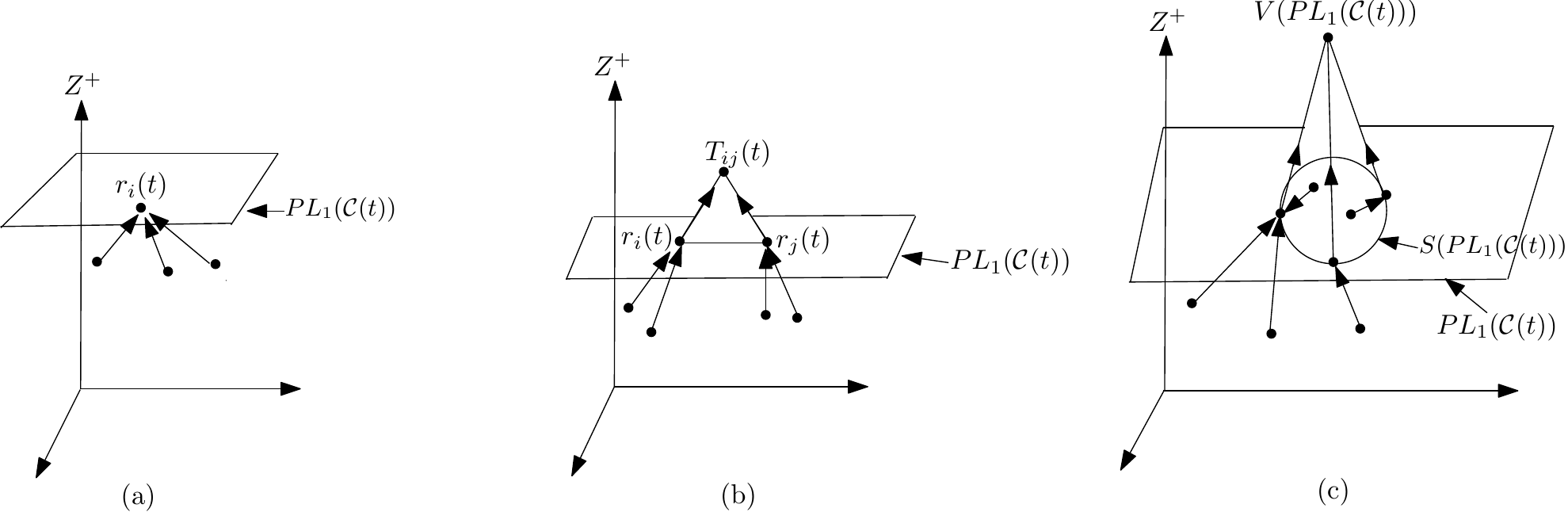}
 \end{center}
 \caption{An example of different scenarios of $Gathering3D$}
 \label{Theo-case-11123} 
\end{figure}
  
 \item {\bf Case 2: $\mathcal C(t) \in \widetilde{\mathcal C_2}$:}
 Let $r_i(t)$ and $r_j(t)$ be two robot positions on $ PL_1(\mathcal C(t))$. The active robots on $ PL_1(\mathcal C(t))$, 
 compute the equilateral triangle $\triangle {r_jT_{ij}r_k}(t)$ and move towards $T_{ij}(t)$ along the  corresponding non-horizontal sides of the triangle. 
A robot which does not lie on the plane $PL_1(\mathcal C(t))$, moves to the nearest point among $r_i(t)$ and $r_j(t)$ (break the tie, if any, arbitrarily).
A robot moves along the straight line joining its current position to the destination point. 

\item {\bf Case 3:  $\mathcal C(t) \in \widetilde{\mathcal C}_{>2}$:}
The robots on $PL_1(\mathcal C(t))$ first compute the circle $S(PL_1(\mathcal C(t)))$. The robots which lie on the circumference of $S(PL_1(\mathcal C(t)))$, compute $W(S(PL_1(\mathcal C(t))),  Z, 45^o)$ and mark $V(PL_1(\mathcal C(t)))$ as their destination point. They move towards $V(PL_1(\mathcal C(t)))$ along the straight lines joining their position to $V(PL_1(\mathcal C(t)))$. 
The robots which lie inside the circle $S(PL_1(\mathcal C(t)))$ and the robots which do not lie on the plane $PL_1(\mathcal C(t))$ move to the respective nearest robot position  on the circumference of $S(PL_1(\mathcal C(t)))$. They move along the 
straight lines joining their current positions to the respective destinations.

Next we present the formal description of the algorithms. The algorithms are executed in all robot sites in their {\it compute} state independently and 
asynchronously. In the main algorithm $Gathering3D$, the robots use $ComputeDestination3D()$ to determine the destination points to move to.

\begin{algorithm}[H]
\KwIn{$ r_i(t), \mathcal P(\mathcal C(t)), RP(\mathcal C(t))$.}
\KwOut{A destination point of $r_i$.}

$h \leftarrow |RPL_1(\mathcal C(t))|$\;
$k\leftarrow CheckLevel3D(\mathcal P(\mathcal C(t)), r_i(t))$\;
\eIf{ $k==1$} 
{
\eIf{ $h==1$}
{$r\leftarrow r_i(t)$\;}
{
\eIf{$h==2$}
{$r \leftarrow ComputeTrianglePeak3D(PL_1(\mathcal C(t)))$\;}
{
$(S(PL_1(\mathcal C(t))), CS(PL_1(\mathcal C(t)))) \leftarrow ComputeCircle3D(PL_1(\mathcal C(t)))$\;
\eIf{$r_i(t)\in CS(PL_1(\mathcal C(t)))$}
{
$r \leftarrow  ComputeConeVertex(S(PL_1(\mathcal C(t))))$\;
}
{$r \leftarrow   ClosestPoint(CS(PL_1(\mathcal C(t))))$\;}
}
}
}
{
$(S(PL_1(\mathcal C(t))), CS(PL_1(\mathcal C(t)))) \leftarrow ComputeCircle3D(PL_1(\mathcal C(t)))$\;
$r \leftarrow   ClosestPoint(CS(PL_1(\mathcal C(t))))$\;
 }
return $r$\;
\caption{ComputeDestination3D()}
\end{algorithm}

\begin{algorithm}[H]
\KwIn{$r_i \in R $}
\KwOut{$r_i$ moves towards its destination.}

Compute $\mathcal P(\mathcal C(t))$\;
Compute $\mathcal RP(\mathcal C(t))$\;
$r \leftarrow ComputeDestination3D(r_i(t), \mathcal P(\mathcal C(t)), RP(\mathcal C(t)))$\;

Move to $r$ along the line segment $\overline{r_i(t)r}$ \;
\caption{Gathering3D()}
\end{algorithm}


\section{Correctness}

In this section, it is established that the gathering of the robots will be achieved, in
finite time, if the robots follow our proposed algorithm. To guarantee gathering in finite time, it has to be shown that a 
point which remains intact under the motion of the robots, can be defined in the system. If the initial configuration admits existence of such a point, then it can 
serve as the gathering point. Otherwise, the movements of the robots are coordinated in such a manner that after a finite time, the initial configuration is changed
to one in which defining such point is possible. Since the robots can not determine multiple occupancy of the robots at a single point, they count all such positions as single points.


%
 \begin{obs}
 \label{equilateral-3D}
  Let $W(S, Z, 45^o)$ be a right circular cone. Let $P=\{P_1, P_2,\ldots,P_k\}$, $k\ge3$, be a set of co-circular points on the surface of $W(S, Z, 45^o)$ such that the plane containing these points is parallel to the base of $W(S, Z, 45^o)$. Let $S'$ be the circle passing through points of $P$. Then the cones $W(S, Z, 45^o)$ and $W(S', Z, 45^o)$ have the same vertex.
 \end{obs}
 
  \begin{lemma}
Suppose $\mathcal C(t_0)) \in \widetilde{\mathcal C}_{2}$ with $r_i(t_0)$ and $r_j(t_0)$ being the two robot positions on $PL_1(\mathcal C(t_0))$. If at least one robot on $PL_1(\mathcal C(t_0))$ starts moving towards $T_{ij}(t_0)$ along the corresponding non-horizontal side of $\triangle{r_iT_{ij}r_j}(t_0)$, then
all the robots on $PL_1(\mathcal C(t))$ will lie on the non-horizontal sides $\triangle{r_iT_{ij}r_j}(t)$ for $t>t_0$.
\end{lemma}
\textbf{Proof}.
 We prove the lemma by induction on the number of completed movements in the system after time $t_0$. Let $l$ denote the number of completed movements in the system. For the base case i.e, for $l=0$, the robots on the topmost plane $PL_1(\mathcal C(t_0))$ lie on two vertices of the the base of $\triangle{r_iT_{ij}r_j}(t_0)$ and the lemma holds.  Suppose the lemma holds up to time $\hat t$, the time when $l^{th}$ movement ends. We prove that the result is also true for $(l+1)^{th}$ movement in the system. Let $t$ be the time  when the $(l+1)^{th}$ movement ends. Now if the $(l+1)^{th}$ movement starts at a time before $\hat t$, then by induction hypothesis, result holds. Otherwise, the robot which makes the $(l+1)^{th}$ movement in the system, must start from point on or below the plane $PL_1(\mathcal C(\hat t))$.
First suppose that a robot not lying on the topmost plane, makes the $(l+1)^{th}$ move. It moves to one of the vertices of the the base of $\triangle{r_iT_{ij}r_j}(t_0)$ and lemma holds. On the other hand, if a robot on the topmost plane, makes the $(l+1)^{th}$ move, it moves towards $T_{ij}(t')$ along the non-horizontal side of $\triangle{r_iT_{ij}r_j}(t')$, where $t_0\le t'<t$. Either it will reach $T_{ij}(t)$ which is same as $T_{ij}(t_0)$ or stops in between.  This implies that the result holds after the $(l+1)^{}th$ 
movement in the system. Since the length of the $(l+1)^{th}$ movement is arbitrary, the result also holds during the motion. 
\qed

 \begin{lemma}
Suppose $\mathcal C(t_0) \in \widetilde{\mathcal C}_{>2}$. Then there exists $t'\ge t_0$, such that all the robots on $PL_1(\mathcal C(t'))$ lie on the surface of $W(S(PL_1 (\mathcal C(t_0))), Z, 45^o)$.
\end{lemma}
\textbf{Proof}. 
Since  $\mathcal C(t_0) \in \widetilde{\mathcal C}_{>2}$, the robots on $PL_1(\mathcal C(t_0))$ compute $S(PL_1 (\mathcal C(t_0)))$ as either the circle of co-circularity   of the robot positions on  $PL_1(\mathcal C(t_0))$ or the minimum 
enclosing circle of the robot positions on  $PL_1(\mathcal C(t_0))$. If $S(PL_1 (\mathcal C(t_0)))$ is the circle of co-circularity, then $t'=t_0$ 
and lemma is true. On the other hand,  if $S(PL_1 (\mathcal C(t_0)))$ is the minimum enclosing circle of the robot positions on  $PL_1(\mathcal C(t_0))$, all 
the active robots on the circumference of $S(PL_1 (\mathcal C(t_0)))$ compute $W(S(PL_1 (\mathcal C(t_0))), Z, 45^o)$ and move to $V(PL_1(\mathcal C(t_0)))$ along the surface of the cone.
Rest of the active robots in the system move towards the nearest robot position on the circumference of $S(PL_1 (\mathcal C(t_0)))$. 
Let $r_i$ be the robot which started moving from the circumference of $S(PL_1 (\mathcal C(t_0)))$ and is the first one to stop. Let  $t'$ be the time when it stops. Since it has been moving along the surface of the cone, its current position is also on the surface of the cone. 
The robots not on the circumference of $S(PL_1 (\mathcal C(t_0)))$, could reach the topmost plane only by reaching the robot positions on circumference of  $S(PL_1 (\mathcal C(t_0)))$.  Irrespective of whether $r_i$ is on the top most plane or not, the only other robots which can lie on the top most plane are the ones which are either going to the vertex of the cone or ones which are following such robots. Hence, $PL_1(C_{t'})$ contains all the robots on surface of the cone. 

\qed

\begin{theorem}
 The algorithm \textit{Gathering-3D} solves the gathering problem in finite time for a set of robots working in three dimensional space under  one-axis agreement with
 arbitrary number of faulty robots.
 
\end{theorem}
\textbf{Proof}. 
Our strategy looks for a point so that all the robots could agree on that point to gather and this point remains intact under the motion of the robots. 
Since robots are oblivious and the scheduling of the actions of the robots are asynchronous, our strategy looks for the invariants present in the robot configurations. If 
 the initial configuration provides such an invariant point, the point serves the purpose. Otherwise, the motion of the robots are coordinated in such a way that after 
 a finite time it would become possible to have such invariants. The algorithm \textit{Gathering-3D}, first classifies the initial configuration $\mathcal C(t_0)$ into any one of the three classes and  then accordingly plans the movements of the robots. 
 \begin{itemize}
  \item {\bf Case 1: $ \boldsymbol{\mathcal C(t_0) \in \widetilde{\mathcal C_1}}$}:
  In this case, $PL_1(\mathcal C(t_0))$ contains a single robot position, say $r_i(t_0)$. According to our algorithm, the robots at $r_i(t_0)$ do not move and rest of the non-faulty robots move towards $r_i(t_0)$ along the straight line joining their respective positions to $r_i(t_0)$. Since no robot reaches a position other than $r_i(t_0)$ 
  on $PL_1(\mathcal C(t_0))$, this point remains intact under the motion of the robots. The point $r_i(t_0)$ serves as the point of gathering.
  
  \item {\bf Case 2: $\boldsymbol{\mathcal C(t_0) \in \widetilde{\mathcal C_2}}$}:
 Here, the plane $PL_1(\mathcal C(t_0))$ contains two robot positions, say $r_i(t_0)$ and $r_j(t_0)$. By lemma 1,  there exists $t'$, such that all the robots on
 $PL_1(\mathcal C(t'))$ will lie on the non-horizontal sides of $\triangle{r_iT_{ij}r_j}(t_0)$ for $t'>t_0$. If $|RPL_1(\mathcal C(t'))|=1$ and all the non-faulty robots are aware of it, then the robot position on $PL_1(\mathcal C(t'))$ can serve our purpose. Otherwise, the robots on $PL_1(\mathcal C(t'))$ compute $T_{ij}(t')$ and move towards it. Once at least one robot reaches $T_{ij}(t')$, the point $T_{ij}(t')$ becomes a static point which can serve as the gathering point. Note that, if for some $t^*\ge t'$,  $|RPL_1(\mathcal C(t^*))|>1$, then  $T_{ij}(t^*)$ and $T_{ij}(t_0)$ are the same point.
 
 \item  {\bf Case 3: $\boldsymbol{\mathcal C(t_0) \in \widetilde{\mathcal C}_{>2}}$}:
 By lemma 2, there exists $t'\ge t_0$, such that all the robots on $PL_1(\mathcal C(t'))$ lie on the surface of $W(S(PL_1 (\mathcal C(t_0))), Z, 45^o)$. If  $|RPL_1(\mathcal C(t'))|\le 2$  and 
 all the non-faulty robots are aware of this, then by Case 1 and Case 2 of the above,  gathering is guaranteed. 
 Otherwise, we analyze the possible scenarios which could occur after the time $t'$ in the execution of our algorithm. Consider an active robot $r_i$ on the plane $PL_1(\mathcal C(t'))$. Due to asynchrony, followings are the possible scenarios for the robot $r_i$:\\
 (i) $r_i$ finds  $|RPL_1(\mathcal C(t'))|=1$ and hence it does not move.\\
 (ii) $|RPL_1(\mathcal C(t'))|=2$. $r_i$ decides to move along a side of an equilateral triangle.\\
 (iii) $|RPL_1(\mathcal C(t'))|>2$. $r_i$ decides to move along the surface of a cone towards the vertex of the cone.\\ 
 (iv) $r_i$ finds itself not on the topmost plane $PL_1(\mathcal C(\hat t))$, $\hat t>t'$ 
 and moves to the nearest robot position on  $S(PL_1(\mathcal C(\hat t)))$.\\
 For scenario (i), $r_i$ decides not to move until any one of the remaining scenarios occurs.
 When scenario (ii) occurs, the robot $r_i$ leaves the surface of $W(S(PL_1 (\mathcal C(t_0))), Z, 45^o)$.  This may also occurs in the scenario (iv). Once robots start leaving the 
 surface of $W(S(PL_1 (\mathcal C(t_0))), Z, 45^o)$, an active robot could find itself in any one of the above four scenarios. It may also find that it does not lie on the circumference of $S(PL_1())$ and will decide to move to the nearest robot position on the same circle $S(PL_1())$. The topmost plane may change many times. To guarantee gathering, we have to show that it will be possible to define an invariant point in finite time. If a robot finds its destination at a distance less than or equal 
 to $\delta$, it reaches there without halting in between. We show that if gathering has not been achieved yet, the configuration will converge to the one in which 
 the topmost plane either contains a single robot position or the geometric span of the robot positions on the topmost plane is at most $\delta$. If former is true then this point will serve the gathering point. For the later case, depending upon the number of robots on the topmost plane, robots directly move to the vertex of cone 
 or to the topmost vertex of an equilateral triangle. Now we measure the maximum decrement in the geometric span of the robot positions on the topmost plane. Let 
 $H(S(PL_1 (\mathcal C(t))), Z)$ be the cylinder with $S(PL_1 (\mathcal C(t)))$ as base and axis parallel to the $Z$ axis. The geometric span of the points on the topmost plane $PL_1 (\mathcal C(t))$ is bounded above by the diameter of 
 $S(PL_1 (\mathcal C(t)))$. When robots on $S(PL_1 (\mathcal C(t)))$  start moving according, the geometric span reduces. Let $a(t)$ denote the radius of $S(PL_1 (\mathcal C(t)))$. 
  \begin{figure}[h]
\begin{center}
 \includegraphics[scale =.45]{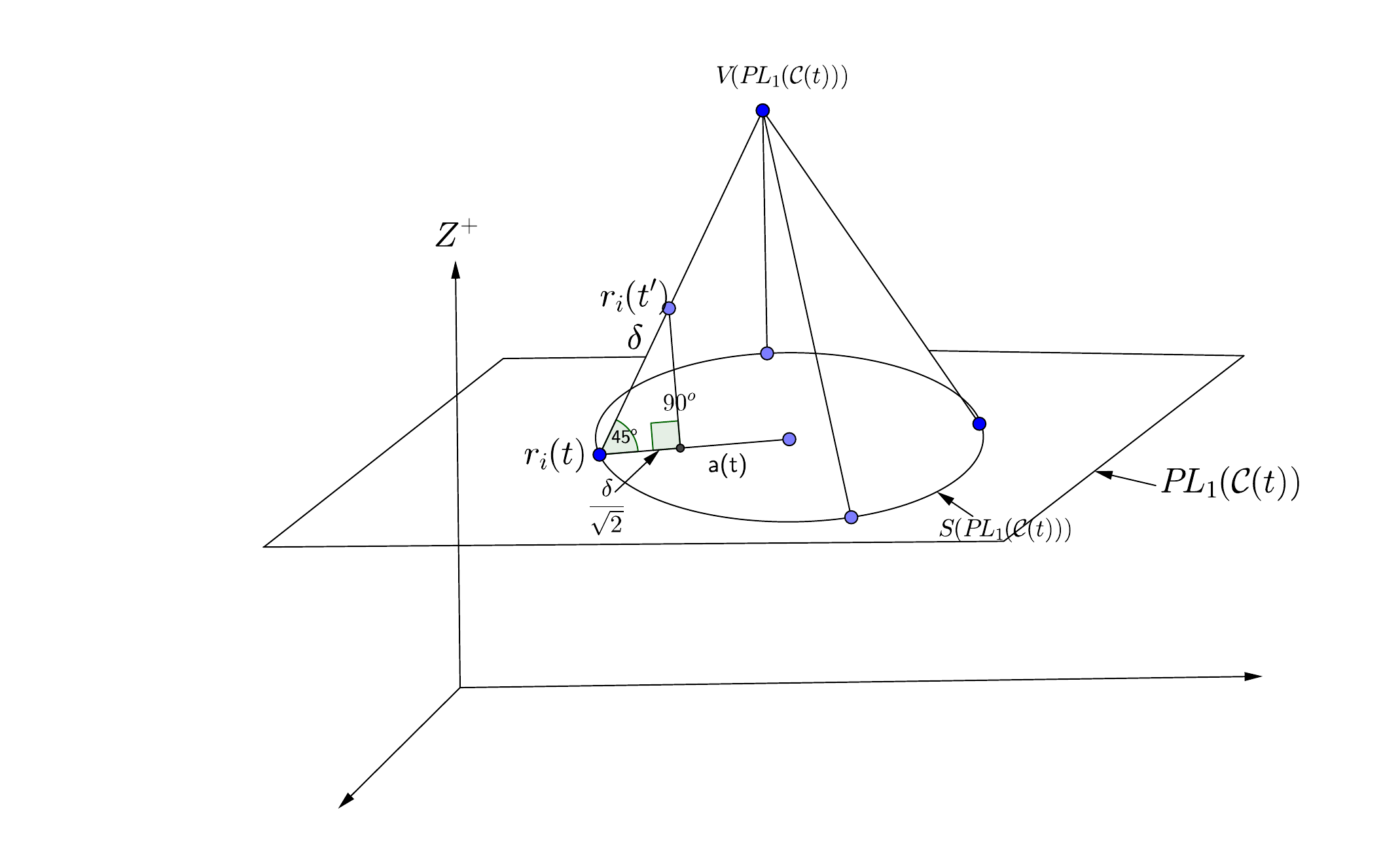}
 \end{center}
 \vspace{-.7cm}
 \caption{An example showing a robot moving along the surface of a cone}
  
 \label{Theo-case-1} 
\end{figure}
 
 There are two scenarios:
\begin{itemize}
 \item \textbf{A robot moves along the surface of a cone:} Let $r_i$ be one of the robots which moves along the surface of the cone $W(S(PL_1 (\mathcal C(t))), Z, 45^o)$.  Suppose  $r_i$ stops at $r_i(t')$. We compute the distance of $r_i(t')$ from the axis of  $W(S(PL_1 (\mathcal C(t))), Z, 45^o)$. If the slant height of the cone is greater than  $\delta$, the distance of $r_i(t')$ from the axis of   $W(S(PL_1(\mathcal C(t))), Z, 45^o)$ is at most $a(t)-\frac{\delta}{\sqrt{2}}$ (Figure \ref{Theo-case-1}). Otherwise, at least one non-faulty active  robot would reach $V(PL_1(\mathcal C_t))$ reducing both $a(t)$ and the number of robot positions on the topmost plane.
 
  \begin{figure}[h]
\begin{center}
 \includegraphics[scale =.6]{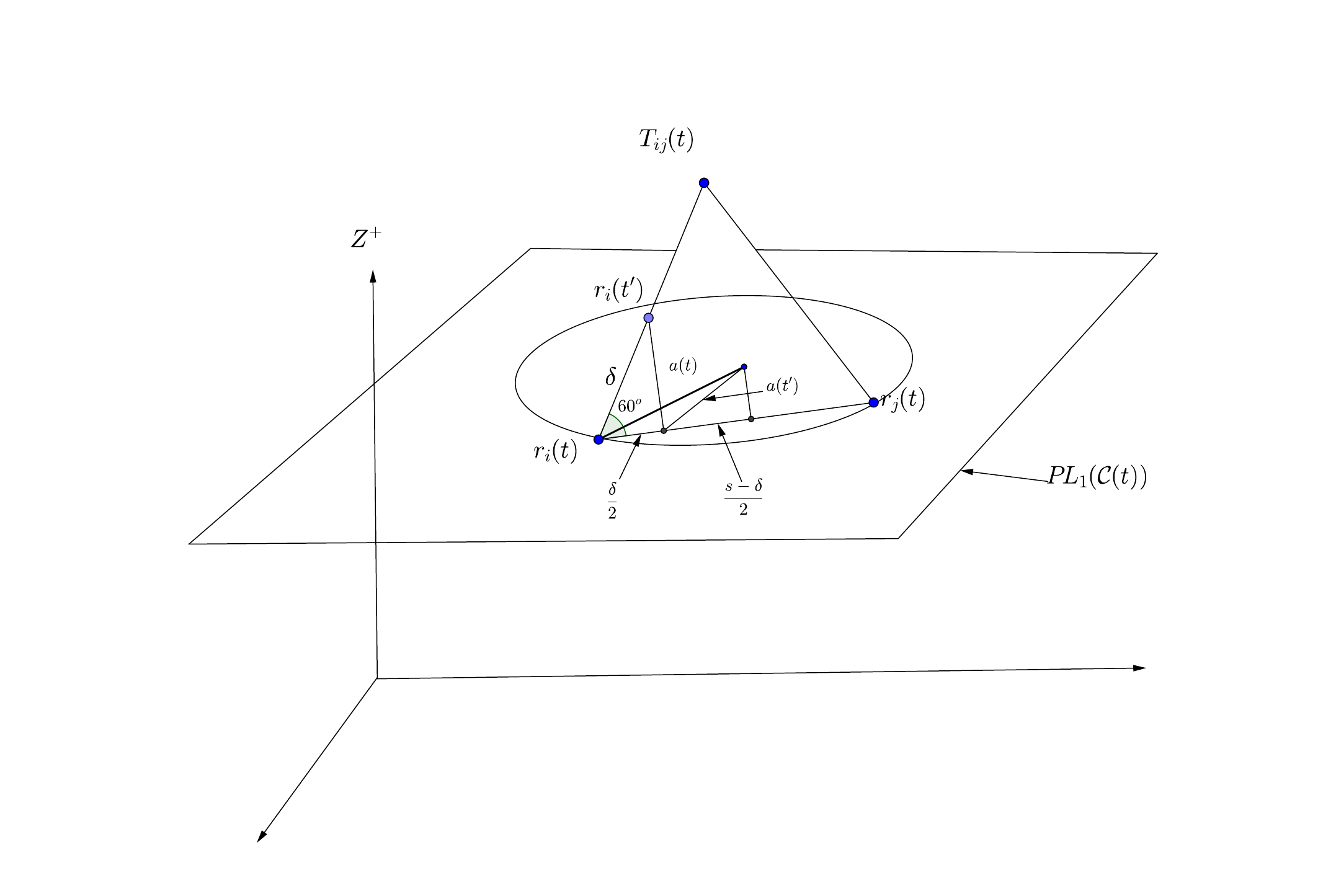}
 \end{center}
 \caption{ An example showing a robot moving along a side of a triangle}
  \vspace{-.5cm}
 \label{Theo-case-2} 
\end{figure}

 \item \textbf{A robot moves along a side of an equilateral triangle:} Let $r_i$ be one of the robots which moves along the side of an equilateral triangle $\triangle r_iT_{ij}r_j(t)$. Suppose $r_i$ stops at $r_i(t')$. If the length of the side of the equilateral triangle is greater than $\delta$, the distance of $r_i(t')$ from the axis of   $W(S(PL_1(\mathcal C(t))), Z, 45^o)$ i.e., the value of $a(t')$ is at most $\sqrt{(a(t))^2-\frac{s^2-(s-\delta)^2}{4}}$, where $s=|\overline{r_i(t)r_j(t)}|$ (Figure \ref{Theo-case-2}). Since $s>\delta$, the reduction in the value of $a(t')$ is bounded below by a function of $\delta$. Otherwise, at least one non-faulty active  robot would reach $T_{ij}(t)$ reducing both $a(t)$ and the number of robot positions on the topmost plane.
\end{itemize}

 Each movement of a robot could reduce the diameter of the base circle of $H(S(PL_1 (), Z)$ by a constant amount. Since the geometric span of the robots in the 
 initial configuration is finite, after a finite number of steps, the geometric span of the robot positions would be reduced down to at most $\delta$. 
 Once the geometric span is reduced to less than or equal to $\delta$, either gathering will be achieved just in next movements of the robots on the topmost plane or the number of robots on the topmost plane will be reduced  and hence finally become one.

 In the proposed algorithm at any step a non-faulty robot always computes a new destination and moves there, unless it is already at the target position for gathering.
  Thus the algorithm can tolerate an arbitrary number of crash faults; the non-faulty robots would still gather at a point.
 \end{itemize}

\qed

 \section{Conclusion}
This paper shows that agreement in one direction is enough for the oblivious, asynchronous robots in 3D space to meet at a single point. Even if some robots become inactive forever, the active robots complete the get-together successfully. The immediate extension of this work would be to consider the opaque robots instead of the transparent point robots and develop a collision-free gathering algorithm for them.  
\end{itemize}

\bibliographystyle{plain}
 \bibliography{Gathering-3D-FUN.bib}

\end{document}